\documentclass[acmtog]{acmart}
\acmSubmissionID{847}

\definecolor{URLBlue}{RGB}{65, 105, 225}
\usepackage{cancel}

\copyrightyear{2024}
\acmYear{2024}
\setcopyright{rightsretained}
\acmConference[SA Conference Papers '24]{SIGGRAPH Asia 2024 Conference Papers}{December 3--6, 2024}{Tokyo, Japan}
\acmBooktitle{SIGGRAPH Asia 2024 Conference Papers (SA Conference Papers '24), December 3--6, 2024, Tokyo, Japan}\acmDOI{10.1145/3680528.3687660}
\acmISBN{979-8-4007-1131-2/24/12}

\begin{document}
\title{Neural Light Spheres for Implicit Image Stitching and View Synthesis}

\author{Ilya Chugunov}
\authornote{Part of this work was done during an internship at Google Inc.}
\email{chugunov@princeton.edu}
\affiliation{
  \institution{Princeton University}
  \country{USA}   
}
\author{Amogh Joshi}
\email{aj0699@princeton.edu}
\affiliation{
  \institution{Princeton University}
  \country{USA}   
}

\author{Kiran Murthy}
\email{murthykk@google.com}
\affiliation{%
 \institution{Google Inc.}
 \country{USA}}
\author{Francois Bleibel}
\email{fbleibel@google.com}
\affiliation{%
 \institution{Google Inc.}
 \country{USA}}
\author{Felix Heide}
\affiliation{%
 \institution{Princeton University}
 \country{USA}}
\email{fheide@princeton.edu}


%



\addtocontents{toc}{\protect\setcounter{tocdepth}{0}}
\begin{abstract}
Challenging to capture, and challenging to display on a cellphone screen, the panorama paradoxically remains both a staple and underused feature of modern mobile camera applications. In this work we address both of these challenges with a spherical neural light field model for implicit panoramic image stitching and re-rendering; able to accommodate for depth parallax, view-dependent lighting, and local scene motion and color changes during capture. Fit during test-time to an arbitrary path panoramic video capture -- vertical, horizontal, random-walk -- these neural light spheres jointly estimate the camera path and a high-resolution scene reconstruction to produce novel wide field-of-view projections of the environment. Our single-layer model avoids expensive volumetric sampling, and decomposes the scene into compact view-dependent ray offset and color components, with a total model size of 80 MB per scene, and real-time (50 FPS) rendering at 1080p resolution. We demonstrate improved reconstruction quality over traditional image stitching and radiance field methods, with significantly higher tolerance to scene motion and non-ideal capture settings.
\end{abstract}

\begin{teaserfigure}
{\includegraphics[width=\textwidth]{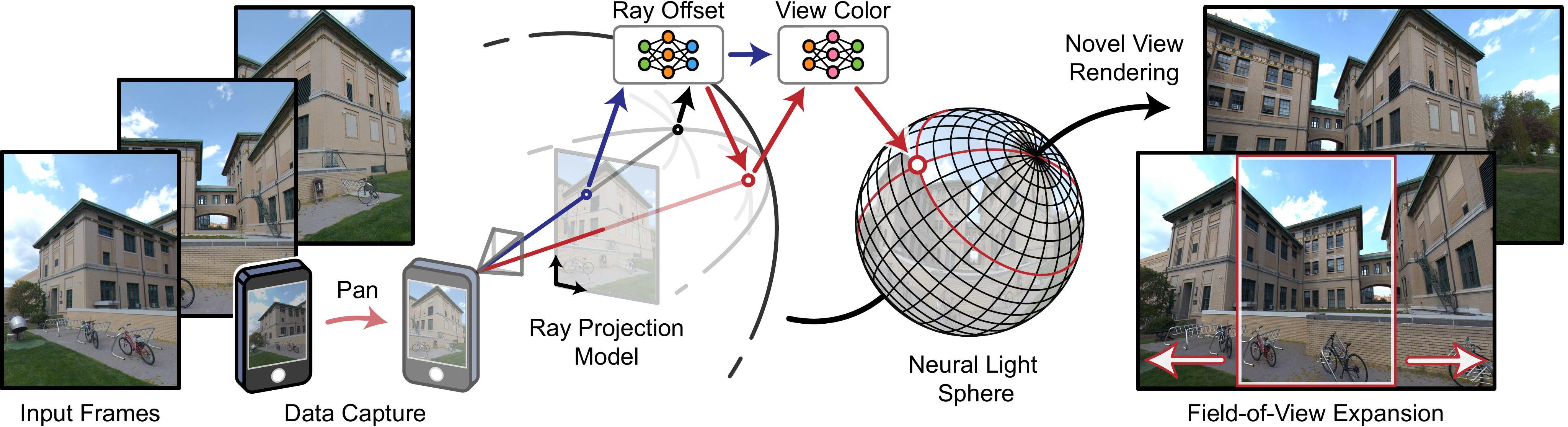}
\caption{Fit during test-time directly to an input panoramic video capture, with no pre-processing steps, our neural light sphere model produces a parallax, lighting, and motion-tolerant reconstruction of the scene. Placing a virtual camera into the sphere, we can generate high-quality wide field-of-view renders of the environment, turning what would otherwise be a static panorama into an interactive viewing experience.}
\label{fig:teaser}}
\end{teaserfigure}

\maketitle

 \vspace{-1em}
\section{Introduction}
The \textit{panorama} of the 19th century was typically a commissioned collection of paintings in a cylindrical arrangement with a dedicated viewing platform to maximize observers' immersion in the work \cite{trumpener2020viewing}. The digital panorama of the 21st century is typically a long rectangle left un-shared -- or un-\textit{viewed} -- in the storage space of the cellphone used to capture it. Yet, arguably the most common form of digital panorama might be the one that is un-\textit{taken}, where the user decides that the hassle of acquisition -- e.g., slowly and carefully sweeping the camera in a level arc across a scene -- is not worth the final product.

To address this imbalance, we can simplify acquisition, increase the appeal of the final product, or (preferably) do both. Moving from cylindrical warping~\cite{szeliski1997creating} and seam matching~\cite{zomet2006seamless} approaches to more parallax-tolerant image stitching processes~\cite{zaragoza2013projective, zhang2014parallax} allows the photographer to take a less restricted camera path and still produce a high-quality panorama. However, the end result remains a single static image. Work on multi-layer depth panoramas~\cite{zheng2007layered, lin2020deep} and panoramic mesh reconstruction~\cite{hedman2017casual} offer a more interactive experience than a traditional panorama, able to use parallax information to render novel views of the scene.  The recent explosion in radiance field methods~\cite{mildenhall2021nerf, kerbl20233d} can be seen as an evolution of ``interactive panoramas'', with a line of connected works from image-based rendering~\cite{chen1993view} to direct view synthesis~\cite{flynn2016deepstereo} and hybrid 3D and image feature approaches~\cite{sitzmann2019deepvoxels}. Neural radiance field (NeRF) methods can produce fast~\cite{muller2022instant} scene reconstructions which model for both parallax and view-dependent lighting effects with high visual quality~\cite{barron2023zip} and from unstructured and unknown poses~\cite{lin2021barf}. However, outward or front-facing panoramas present a major challenge for these volumetric representations, as large parts of the scene are only observed for a few frames before falling out of view, turning scene reconstruction into a collection of sparse view problems~\cite{niemeyer2022regnerf}.

In this work we explore a compact neural light field~\cite{attal2022learning} model for panoramic image stitching and view synthesis; capable of encoding depth parallax, view-dependent lighting, and local scene motion and color changes. We represent the scene as a color-on-a-sphere model decomposed into two components: a view-dependent ray offset model for parallax, lens distortion, and smooth motion; and a view-dependent color model for occluded content, reflections, refraction, and color changes. Taking as input an arbitrary path panorama -- vertical, horizontal, random-walk -- we fit our model at \textit{test-time} to jointly estimate the camera path, and produce a high-resolution stitched representation of the scene. We demonstrate how this model enables geometrically consistent field-of-view expansion, transforming portrait-mode panoramas into immersive, explorable wide-view renders.

Specifically, we make the following contributions:
\begin{itemize}
  \item A compact and efficient (80 MB model size per scene, 50 FPS rendering at 1080p resolution) two-stage neural light sphere model of panoramic photography.
  \item Validation of panoramic image stitching and view synthesis performance under varying imaging settings, including low-light conditions, with comparisons to traditional image stitching and radiance field approaches. 
  \item An Android-based data collection tool for streaming and recording full-resolution RAW image arrays, camera and system metadata, and on-board device measurements such as gyroscope and accelerometer values.
  \item A diverse collection of 50 indoor and outdoor handheld panoramic scenes recorded from all three on-device cameras with full 10-bit color depth, 12-megapixel resolution.
\end{itemize}

\noindent We make our code, data, and data collection app available open-source on our project website: {\color{URLBlue}\href{https://light.princeton.edu/NeuLS}{light.princeton.edu/NeuLS}}


\section{Related Work}
\paragraph{Image Stitching} There is a rich history of methods for stitching or \textit{mosaicing}~\cite{burt1983multiresolution} multiple images into one, with demand for the task long pre-dating the invention of digital photography~\cite{shepherd1925interpretation}. A common approach is to first extract image features, either directly calculated~\cite{brown2007automatic, lowe2004distinctive} or learned~\cite{sarlin2020superglue}, which are matched to position and warp images together~\cite{gao2011constructing}. Allowing for image transforms beyond simple homographies~\cite{hartley2003multiple} can allow for parallax-tolerant image warping and stitching~\cite{shum2002construction, zhang2014parallax}, reducing blur from pixel disparity between views. Seam-carving approaches dynamically adjust the stitching boundaries to better match visual features~\cite{agarwala2004interactive, gao2013seam}, helping to avoid artifacts from mismatched content on image boundaries. Inspired by local deformation image stitching~\cite{zaragoza2013projective} and panoramic video texture~\cite{agarwala2005panoramic} work, we develop a neural field model which can accommodate for both parallax and scene motion during reconstruction. However, rather than use sparse pre-computed features and break the reconstruction pipeline into multiple discrete steps, we leverage a neural scene representation and fast ray sampling to optimize our model end-to-end over dense pixel-wise photometric loss.

\begin{figure*}[t!]
 \centering
\includegraphics[width=\textwidth]{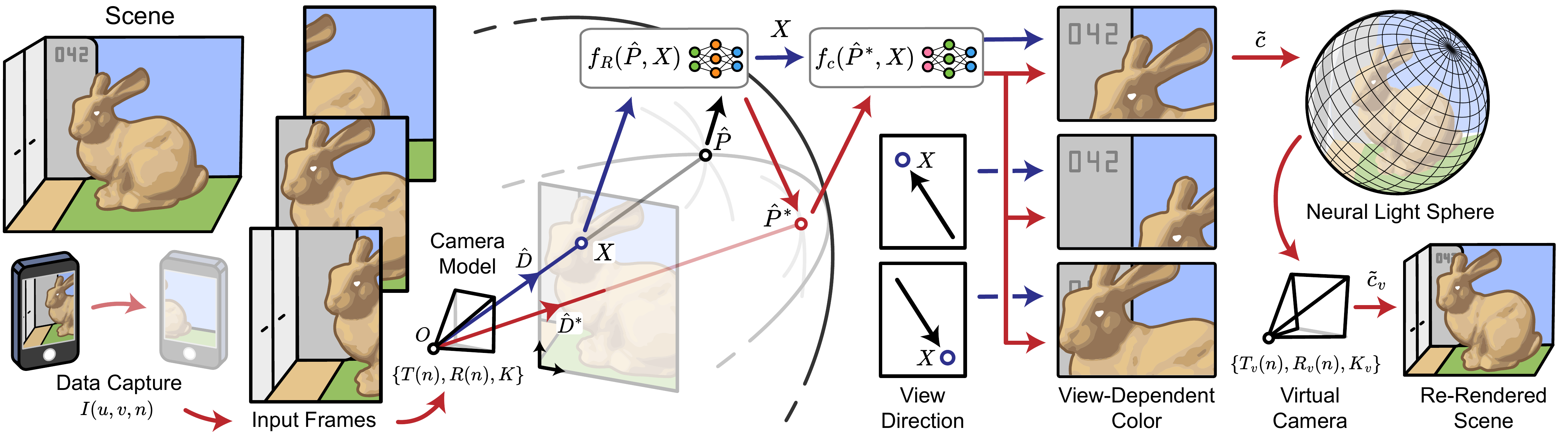} \vspace{-2em}
  \caption{\textbf{Neural Light Sphere Model.} Taking as input panoramic video capture $I(u,v,n)$, we perform backward camera projection from a point $X=(u,v)$ into a spherical hull to estimate an initial intersection point $P$. Ray offset model $f_{\textsc{r}}(\hat{P},X)$ then bends this ray to a corrected point $\hat{P}^*$, which is used to sample the view-dependent color model $f_{\textsc{c}}(\hat{P}^*,X)$. Simulating a new virtual camera with our desired position and FOV, we use this neural light sphere model to re-render the scene to novel views.} \vspace{-1em}
\label{fig:method}
\end{figure*}

\paragraph{Layered and Depth Panoramas} Concentric mosaic~\cite{shum1999rendering} and layered depth map~\cite{shade1998layered} representations offer a compact way to model the effects of parallax and occlusion in a scene. Layered depth panoramas~\cite{zheng2007layered} make use of a layered representation to produce an interactive image stitching reconstruction, able to render novel views through trigonometric reprojection. Follow-on work extends this reconstruction to mesh representations~\cite{hedman2017casual,hedman2018instant} and learned features~\cite{lin2020deep}, offering improved reconstruction of object surfaces which are otherwise occluded between depth layers. Work in this space often targets VR applications~\cite{bertel2020omniphotos, lai2019real, attal2020matryodshka}, as they drive demand for high-quality immersive and interactive user experiences in 3D environments. Also related are video mosaic approaches~\cite{rav2008unwrap, kasten2021layered}, which forgo re-rendering to decompose a video into a direct 2D-to-2D pixel mapping onto a set of editable atlases. In this work, we target reconstructions that can provide an interactive user experience with minimal hardware or camera motion requirements~\cite{bertel2020omniphotos}, and which are able to tolerate moderate scene motion and color changes.

\paragraph{Light Field Methods} Modeling ray color as a product of three dimensional spatial and angular components, a light field can fully represent effects of depth parallax, reflections, and refraction in a scene~\cite{lfr,ng2005light} at the cost of high data, storage, and computational requirements~\cite{wilburn2005high}. Lumigraphs~\cite{gortler1996lumigraph} make use of a simpler geometric proxy -- e.g., the crossing points of a ray intersecting with two planes -- to represent the spatial and angular components of a light field, greatly lowering data and computational requirements for reconstruction and rendering~\cite{chai2000plenoptic}. Motivated by recent work in neural light field representations~\cite{attal2022learning,suhail2022light}, we develop a compact spherical representation which decomposes the scene into view-dependent ray offset -- for effects such as parallax and local motion -- and view-dependent color for occlusions and time-dependent content.
\paragraph{Neural Scene Representations} Recent work in neural scene representations, particularly in the area of neural radiance fields (NeRFs) \cite{barron2023zip,mildenhall2021nerf}, has demonstrated that high quality scene reconstruction can be achieved without pixel arrays, voxel grids, or other explicit backing representations. These approaches train a neural network at \textit{test time} -- starting with an untrained network, overfit to a single scene -- to map from encoded~\cite{tancik2020fourier} coordinates to output parameters such as color~\cite{nam2022neural}, opacity~\cite{martin2021nerf}, density~\cite{corona2022mednerf}, depth~\cite{chugunov2023shakes}, camera lens parameters~\cite{xian2023neural}, and surface maps~\cite{morreale2021neural}. While they are not neural scene representations, forward projection ``Gaussian Splatting''~\cite{kerbl20233d} models have recently exploded in popularity as an alternative to NeRF scene representations, offering increased rendering speed by avoiding costly volume sampling operations. However, outward panoramic captures with largely rotational motion present a challenge for these methods, which rely on large view disparity to localize content in 3D space. We instead propose a view-dependent ray offset and color model to reconstruct local parallax and view-dependent effects from minimal view disparity. By embedding this representation on a spherical surface, we also substitute costly NeRF volume sampling with efficient ray-sphere crossings, resulting in a compact 80 MB model capable of real-time 1920x1080px rendering at 50 FPS.
\section{Neural Light Sphere Reconstruction}
In this section we describe our proposed neural light sphere model for implicit image stitching and re-rendering. We begin with an overview of our backward projection model for unstructured panoramic captures. We then discuss the neural field representations backing this model, its loss and training procedure, how we collect scene data for reconstruction, and implementation details.

\subsection{Projective Model of Panoramic Imaging}
\label{sec:projective_model}
In this work, we adopt a spherical backward projection model~\cite{szeliski2007image} for our scene representation. That is, we model each image in the input video as the product of rays originating at the camera center intersecting with the inner surface of a sphere. To simplify notation, we outline this process for a single ray below, illustrated in Fig.~\ref{fig:method}, and later generalize to batches of rays. Let
\begin{align}
    c = [\mathrm{R},\mathrm{G},\mathrm{B}]^\top = I(u,v,n)
\end{align}
be a colored point sampled at image coordinates $u,v\in [0,1]$ from a frame $n \in [0,N{-}1]$ in a video $I(u,v,n)$, where $N$ is the total number of captured video frames. To project this point to a camera ray, we introduce camera rotation $R(n)$ and translation $T(n)$ models
\begin{align}\label{eq:translation_rotation}
    &T(n) = \mathbf{T}_{n}, \quad R(n) = \mathrm{rot}(\eta_\textsc{r}\mathbf{R}_n) \mathbf{G}_n \nonumber \\
    &\mathbf{T}_{n} =  
    \left[\arraycolsep=2.0pt
    \begin{array}{c}
    t_x \\
    t_y \\
    t_z \\
    \end{array}\right],\,\,
    \mathbf{R}_{n} =  
    \left[\arraycolsep=2.0pt
    \begin{array}{c}
    r_x \\
    r_y \\
    r_z \\
    \end{array}\right],\,\,
    \mathrm{rot}(\mathbf{R}_n)  = \left[\begin{array}{ccc}
1 & -r_z & r_y \\
r_z & 1 & -r_x \\
-r_y & r_x & 1
\end{array}\right].
\end{align}
Here, we model translation for frame $n$ as three dimensional motion, initialized at zero. $R(n)$ is a small-angle approximation~\cite{boas2006mathematical} offset $\mathbf{R}_n$ to device rotation $\mathbf{G}_n$ recorded from the phone onboard gyroscope, weighted by $\eta_\textsc{r}$. With calibrated intrinsics matrix $K$, sourced from device camera metadata, we project the point at $u,v$ sampled from frame $n$ to a ray with origin $O$ and direction $D$ as
\begin{align}\label{eq:ray_generation}
    O \,{=}\, \left[\arraycolsep=2.0pt
    \begin{array}{c}
    O_x \\
    O_y \\
    O_z \\
    \end{array}\right] = T(n), \, \, \quad D = \left[\arraycolsep=2.0pt
    \begin{array}{c}
    D_x \\
    D_y \\
    D_z \\
    \end{array}\right] = R(t)K^{-1} \left[\arraycolsep=2.0pt
    \begin{array}{c}
    u \\
    v \\
    1 \\
    \end{array}\right].
\end{align}
We normalize the direction vector $\hat{D} = D/\left\|D\right\|$ to simplify reprojection steps. Next, we define our image model to lie on the surface of a sphere, and calculate its intersection point $P$ with this ray as
\begin{align}\label{eq:sphere_intersection}
   \hat{P} &=  P/\left\|P\right\|, \quad P \,{=}\, \left[\arraycolsep=2.0pt
    \begin{array}{c}
    P_x \\
    P_y \\
    P_z \\
    \end{array}\right] = O + t\hat{D}\nonumber\\
    t &= -\left(O \cdot \hat{D}\right) + \sqrt{(O \cdot \hat{D})^2 - (\left\| O\right\|^2 - 1)},
\end{align}
assuming a sphere of radius 1, centered at $[0,0,0]^{\top}$, with the ray originating within its radius ($\left\| O\right\|^2 < 1$). However, as this sphere model, in general, does not match the true scene geometry, we introduce a ray offset model $f_{\textsc{r}}(\hat{P},X)$ to offset the ray direction as

\begin{align}\label{eq:ray_distortion}
   \hat{D}^* =  D^*/\left\|D^*\right\|, \quad D^* = \mathrm{rot}\left(\mathbf{R} = f_\textsc{r}(\hat{P}, X)\right)\hat{D},
\end{align}
where $X = [u,v]^\top$ is the ray's originating image coordinates, and $\mathrm{rot(\mathbf{R})}$ is the small-angle rotation model from Eq.~\ref{eq:translation_rotation}. We can observe that this model generalizes effects such as parallax (deflecting rays as a function of position via $\hat{P}$) and lens distortion (deflecting rays as a function of their angle relative to the camera center via $X$). With this corrected ray $(O, \hat{D}^*)$, we re-sample our sphere via Eq.~\ref{eq:sphere_intersection} to generate a new intersection point $\hat{P}^*$. To map this point to estimated scene color $\tilde{c}$, we introduce a view-dependent color model $f_\textsc{c}$, where
\begin{align}\label{eq:ray_distortion}
   \tilde{c} = [\tilde{\mathrm{R}},\tilde{\mathrm{G}},\tilde{\mathrm{B}}]^\top = f_\textsc{c}(\hat{P}^*, X).
\end{align}
This model takes as input the camera coordinate $X$, which allows for modeling of view-dependent effects such as occlusions, reflections, motion, and generated content (e.g., flashing lights), and maps it together with the ray intersection on the sphere $\hat{P}^*$ to an output estimated RGB value $\tilde{c}$. To generate novel views we take as input virtual camera intrinsics $K_v$, translation $T_v(n)$ and rotation $R_v(n)$, and repeat Eq.~\eqref{eq:translation_rotation}--\eqref{eq:ray_distortion} with these new parameters to generate a colored point $\tilde{c}_v$.
\begin{figure}[t!]
 \centering
\includegraphics[width=\linewidth]{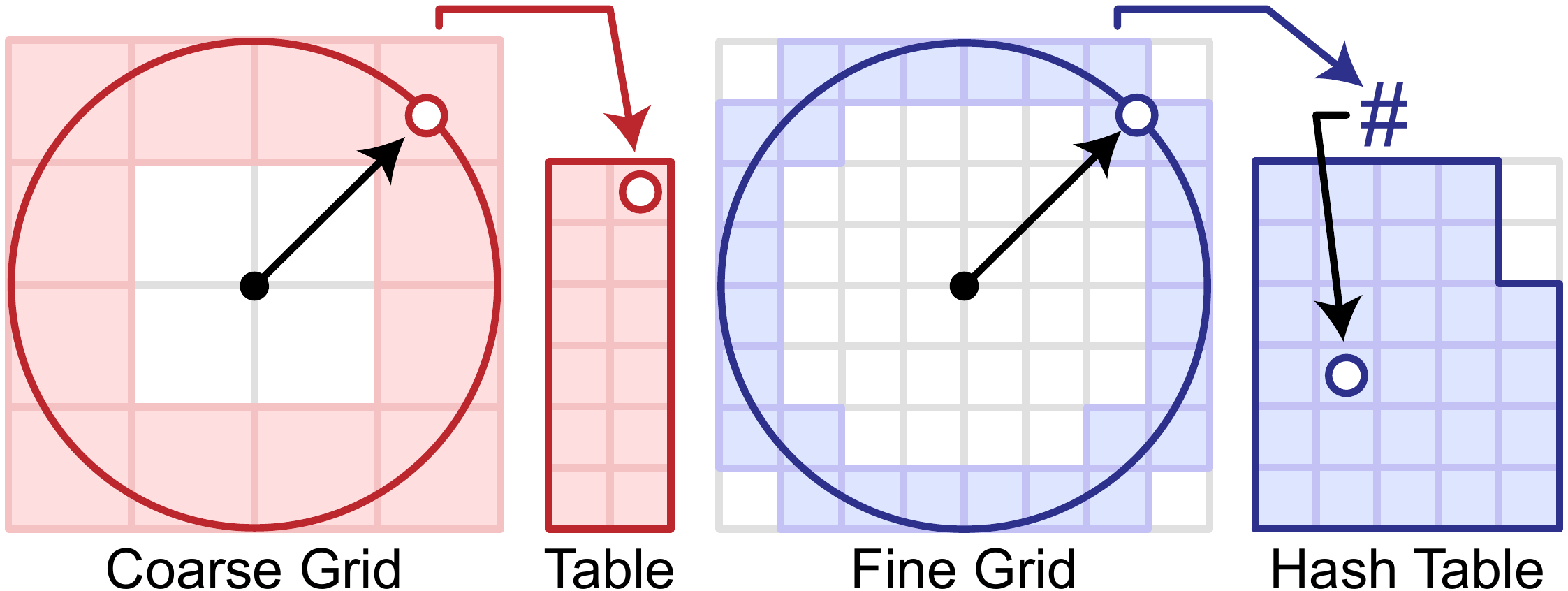}
  \caption{\textbf{Hash Grid Spheres.} In this 2D example we can observe how, for points on a circle, the number of accessed elements in the backing grid roughly doubles for a squaring of grid elements. Given an efficient mapping from grid location to element -- e.g., hash table lookup -- this forms a compact representation even at high resolutions, where storing a dense grid would be computationally intractable.} \vspace{-1em}
\label{fig:hash_sphere}
\end{figure}

\subsection{Neural Field Representations} \label{sec:neural_fields}
In the section above, we introduce, but do not \textit{define}, our two core models: $f_\textsc{r}$ for ray offset, and $f_c$ for view-dependent color estimation. Much of the diversity in image stitching and view synthesis approaches can be seen as design choices for these models. For example, $f_\textsc{r}$ could be a layered depth model~\cite{shade1998layered} or cylindrical projection~\cite{mcmillan1995plenoptic}, and $f_\textsc{c}$ could be an explicit color blending~\cite{buehler2001unstructured} or implicit radiance field~\cite{mildenhall2021nerf}, each with tradeoffs in representation power, extrapolation, and input data requirements. With this in mind, we aim to design $f_\textsc{r}$ and $f_\textsc{c}$ to produce a system which is:
\begin{enumerate}
    \item \textbf{Compact}: such that that model is simple to train and has low memory and disk space usage. Thus we minimize the number of components, networks, loss and regularization functions, and avoid pre-processing steps (such as COLMAP~\cite{schonberger2016structure}).
    \item \textbf{Robust}: able to reconstruct a wide range of capture settings (indoor, outdoor, night-time), capture paths, and scene dynamics (e.g., moving clouds, blinking lights). Failing gracefully for hard-to-model effects, with localized reconstruction errors.
\end{enumerate}
\begin{figure}[t!]
 \centering
    \includegraphics[width=\linewidth]{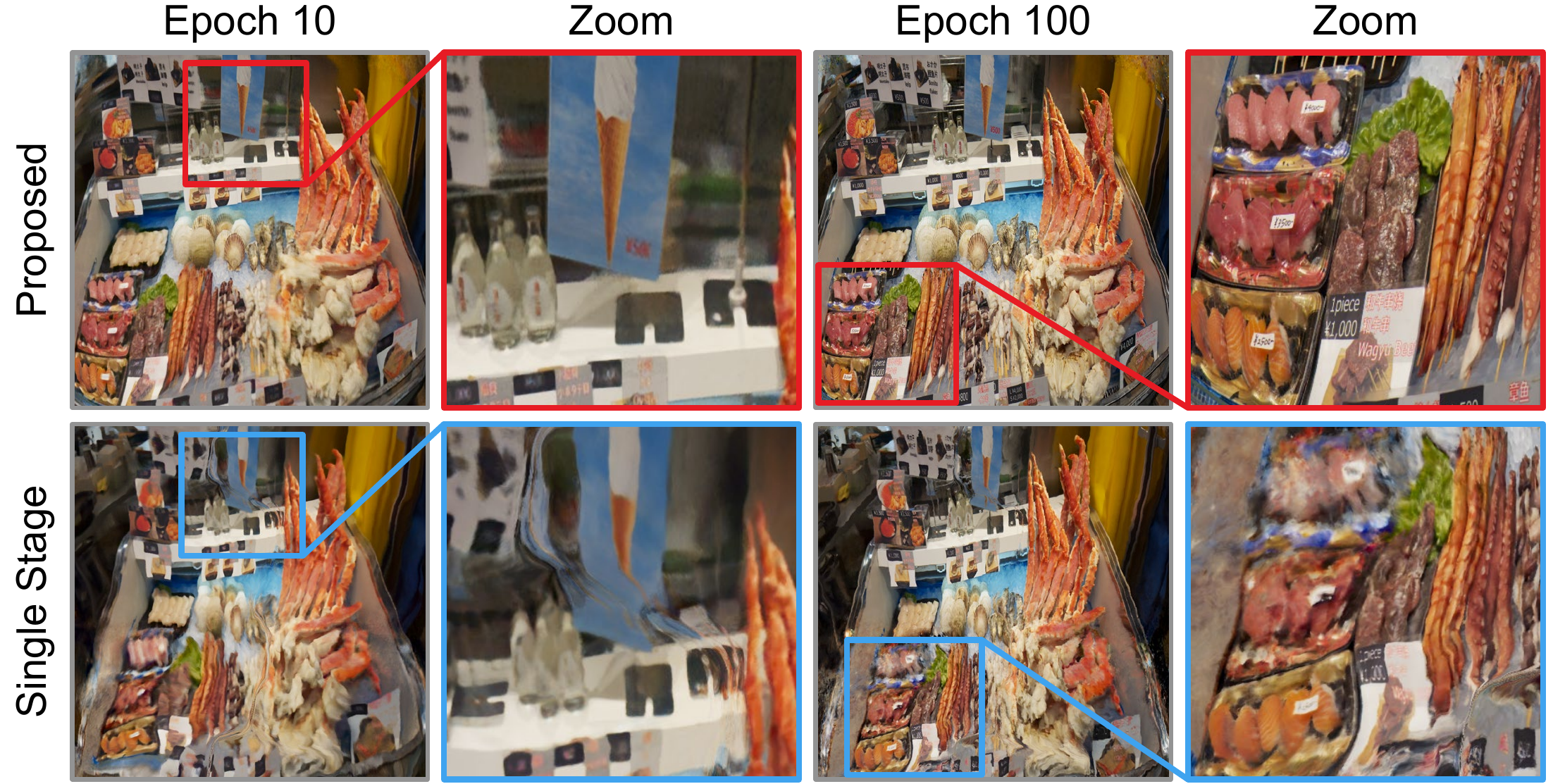}
  \caption{\textbf{Two Stage Training.} Breaking training into two stages allows the camera pose and static image model to first fit an approximation of the scene before view-dependent effects are introduced via $h_\textsc{r}$ and $h_\textsc{d}$. This helps avoid artifacts during early training, like the discontinuities around the sign in the \textit{Single Stage} example, which result in poor final reconstruction quality.} \vspace{-1em}
\label{fig:two_stage}
\end{figure}
Neural scene representations, particularly with high-level hardware-optimized implementations~\cite{muller2022instant}, offer compelling solutions to this design challenge. By implicitly representing the scene in the weights of a multi-layer perceptron (MLP)~\cite{hornik1989multilayer}, we can effectively turn data storage and retrieval into a component of our inverse imaging model. Correspondingly, we represent ray offset $f_\textsc{r}$ as
\begin{align}
    f_\textsc{r}(\hat{P}, X) = h_\textsc{r}(\gamma_1 (\hat{P}) \oplus \gamma_1 (X);\, \theta_\textsc{r}),
\end{align}
where $\oplus$ denotes concatenation. Here, $h_\textsc{r}$ is an MLP with learned weights $\theta_\textsc{r}$, and $\gamma_1$ is the multi-resolution hash grid encoding from \cite{muller2022instant}, sampled with 3D normalized ray intersection $\hat{P}$ and 2D camera coordinate $X$. During training, $h_\textsc{r}$ learns a mapping between these encoded vectors and the offset applied to $\hat{D}\rightarrow\hat{D}^*$. We similarly construct the view-dependent color model $f_\textsc{c}$ as
\begin{align}
    f_\textsc{c}(\hat{P}^*, X) = h_\textsc{c}\left(h_\textsc{p}(\gamma_2 (\hat{P}^*); \,\theta_\textsc{p}) + h_\textsc{d}(\gamma_1 (X);\, \theta_\textsc{d}); \, \theta_\textsc{c}\right),
\end{align}
The network $h_\textsc{d}$ takes as input camera coordinate $X$ and outputs a vector encoding of view direction; network  $h_\textsc{p}$ similarly encodes the corrected position of the sphere crossing. This combined encoding is then mapped to color via $h_\textsc{c}$. Of note is that $\gamma_2$, the multi-resolution hash encoding applied to $\hat{P}$, and $\gamma_1$, the encoding applied to $\hat{P}^*$, operate in 3D world space on the surface of the unit sphere. That is, \textit{we never convert intersections to spherical coordinates}, and avoid the associated non-linear projection~\cite{zelnik2005squaring} and singularity problems. While it would be exceedingly inefficient to store a sphere in a dense representation of sufficient resolution for high-quality image synthesis (e.g., $4000^3$ voxels for 12-megapixels images, the majority of which would be empty), this is made possible thanks to the hash-grid backing of $\gamma$. Illustrated in Fig.~\ref{fig:hash_sphere}, as the majority of the grid locations inside in the unit cube are never sampled, since they do not intersect with the unit sphere's surface, the corresponding stored entries in $\gamma$ are never queried. Thus the size of the hash table for $\gamma$ -- which determines its latency, memory usage, and storage requirements -- can be on the order of magnitude of the sphere's surface area rather than its volume.

\begin{figure}[t!]
 \centering
\includegraphics[width=\linewidth]{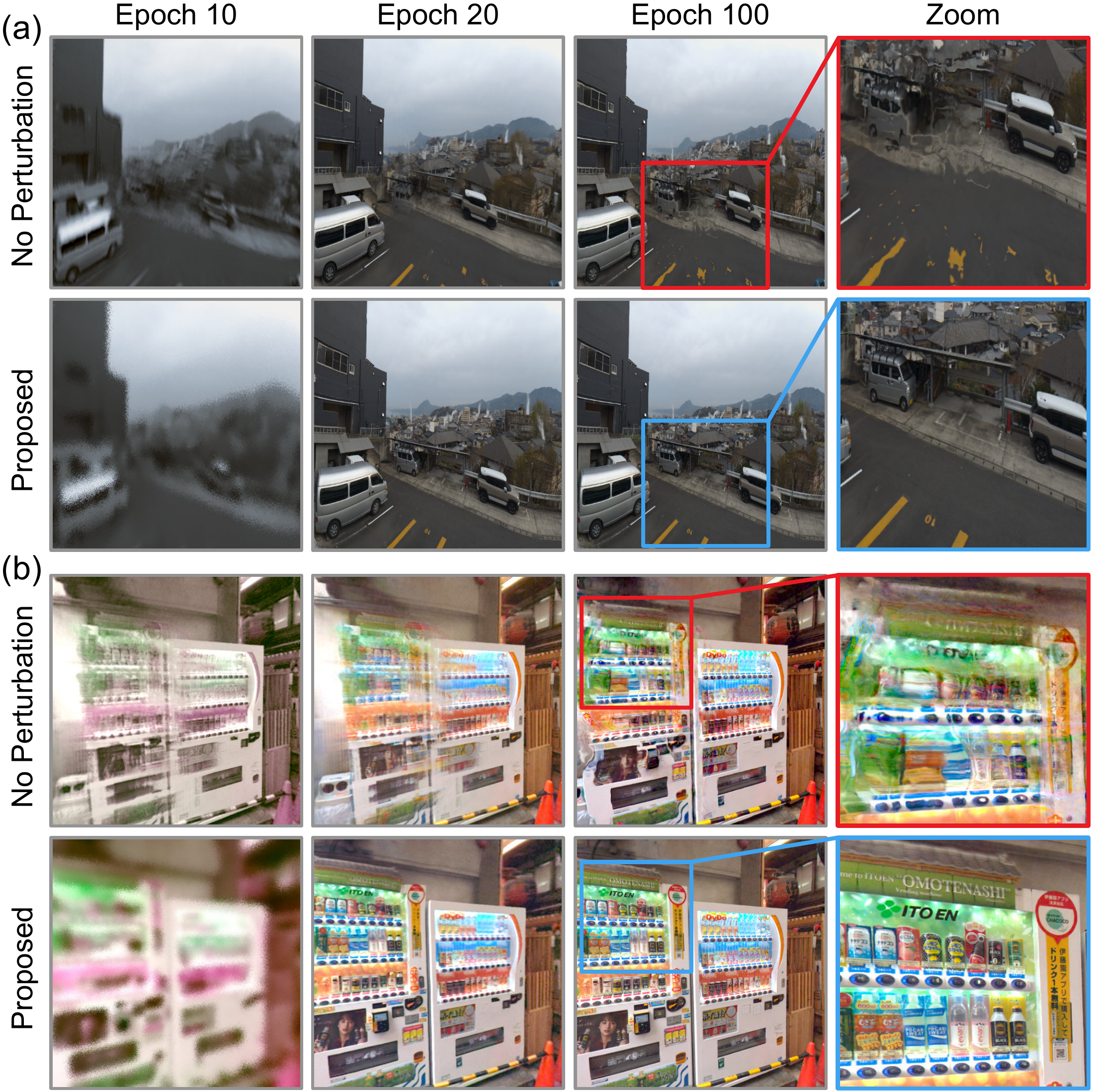}
  \caption{\textbf{Ray Perturbations.} By applying small perturbations to ray origins $O$ we are able to avoid hard-to-escape local minima solutions during early training epochs. In (a) we see how for the road, a region with low image texture, the \textit{No Perturbation} example duplicates content; creating two copies of the \#10 parking spot. In (b) we see how for repeated textures, perturbations can also help avoid ``crunching'' content in early training, where the repeated cans in the vending machine are accidentally aligned on top of each other. }
\label{fig:perturbations}
\end{figure}

\subsection{Loss and Training Procedure}
With the rotation model $R(n)$ initialized with the device's onboard gyroscope measurements, and the translation model $T(n)$ initialized as all zeroes, we train the networks $\{h_\textsc{r}, h_\textsc{c}, h_\textsc{p}, h_\textsc{d}\}$ from scratch via stochastic gradient descent to fit an input scene. We break training into two stages: in the first, we freeze the ray offset and view-dependent color networks $h_\textsc{r}, h_\textsc{d}$ to allow the model to learn initial camera pose estimates and spherical color map, and in the second stage we unlock all networks to let them jointly continue training. Illustrated in Fig.~\ref{fig:two_stage}, this helps prevent image artifacts caused by $h_\textsc{r}, h_\textsc{d}$ from accumulating during early training, where it is uncertain if parts of the scene are undergoing view-dependent color changes or simply stereo parallax. A similar problem also occurs for training the sphere color networks $h_\textsc{c}, h_\textsc{p}$, where the multi-resolution hash encoding $\gamma$ allows the network to fit image content \textit{undesirably} fast. This leads to artifacts, as seen in Fig.~\ref{fig:perturbations}, where the image model learns duplicated or overlapping content faster than the motion model can correct for. We find that an effective and computationally inexpensive way of combating this behavior, shown in Eq.~\eqref{eq:perturbation}, is to add small perturbations to rays generated via Eq.~\eqref{eq:ray_generation} as
\begin{align}\label{eq:perturbation}
    \tilde{O} = O + \eta_p\mathcal{N}(0, 1),
\end{align}
where $\mathcal{N}(0, 1)$ is zero-mean standard Gaussian noise. The weight term $\eta_p$ is gradually decayed to zero over the first stage of training. Similar to prior work~\cite{chugunov2023shakes, li2023neuralangelo} we also mask the highest frequency grids in $\gamma_1$ and $\gamma_2$ to reduce the amount of accumulated noise during early training.

Given linear RAW inputs, we find $L_1$ to be an effective training loss, particularly for high noise reconstruction where zero-mean Gaussian read noise~\cite{brooks2019unprocessing} can be averaged out:
\begin{align}\label{eq:loss}
    \mathcal{L} = |c - \tilde{c}|.
\end{align}
We find that, with careful selection of encoding parameters for $\gamma_1$ and $\gamma_2$, \emph{no additional explicit regularization penalties are required} to constrain scene reconstruction~\cite{chugunov2024neural}. 

\begin{figure}[t!]
 \centering
\includegraphics[width=\linewidth]{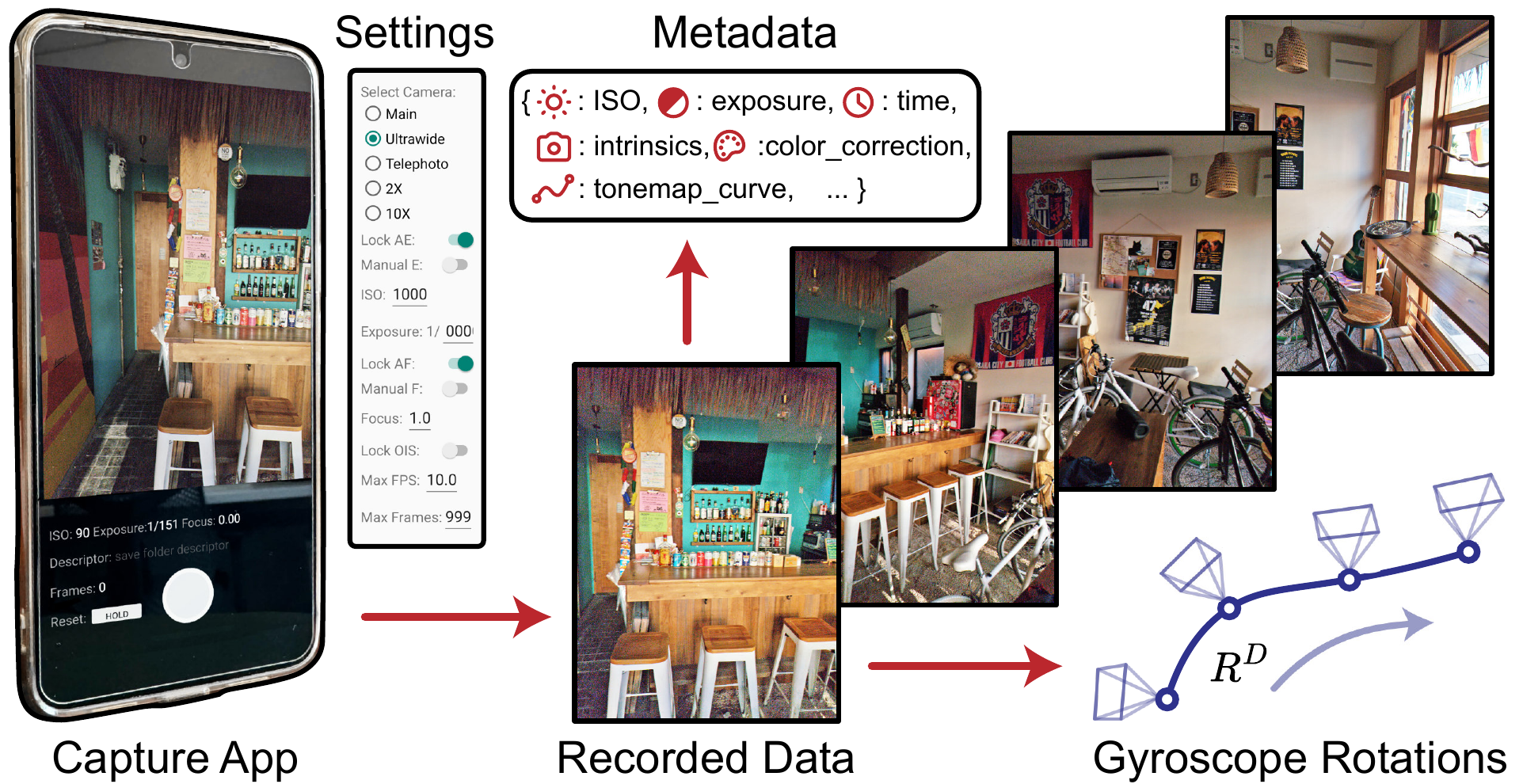}
  \caption{\textbf{Data Capture.} We develop an open-source Android-based mobile application to facilitate in-the-wild capture of scenes. The app's settings allow for camera selection (main, ultrawide, or telephoto) and to either use the device's auto-focus and auto-exposure features for capture, or set their respective values. During capture, we record full resolution Bayer RAW images, device accelerometer and gyroscope measurements, and all exposed camera and frame metadata including: ISO, exposure, timestamps, camera intrinsics, and color and tone correction values. }
\label{fig:app_layout}
\end{figure}
\begin{figure*}[t!]
 \centering
\includegraphics[width=\textwidth]{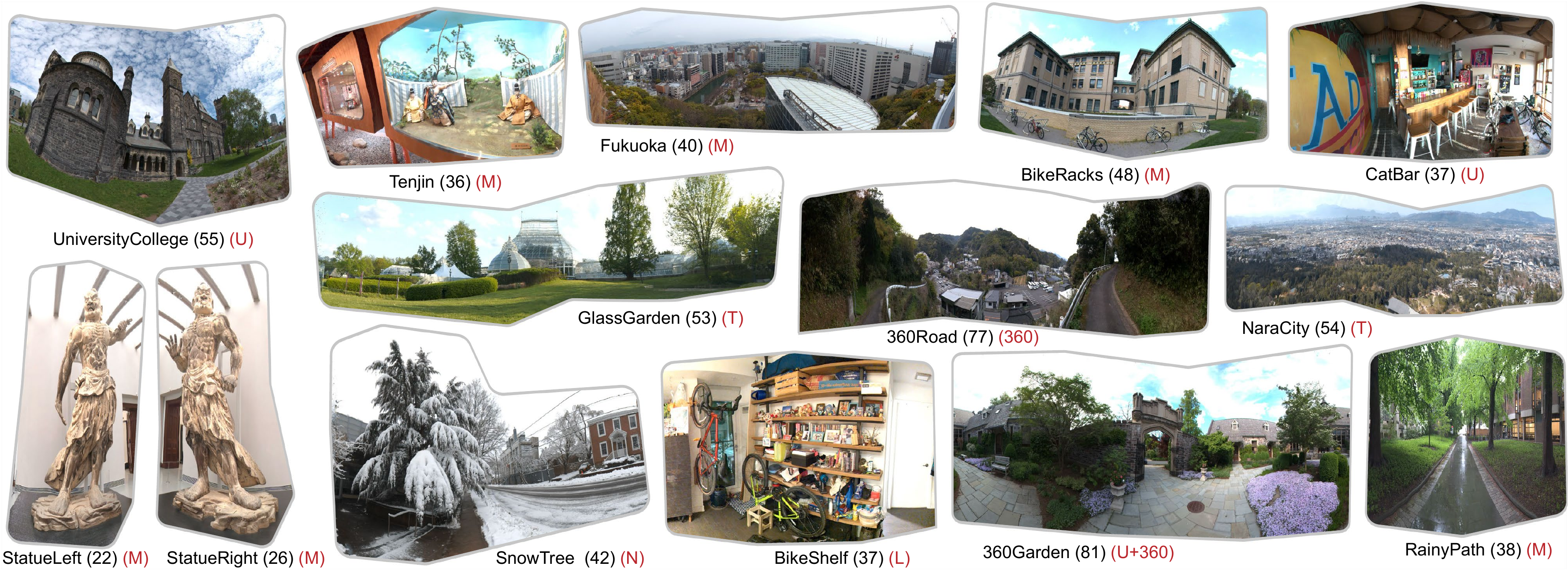}
  \caption{\textbf{Scene Diversity.} Shown above are spherical re-projections of reconstructions for a representative subset of scenes from our collected dataset. These include: (M) 1x main lens, (U) 0.5x ultrawide, (T) 5x telephoto, (L) low-light, (N) non-linear, and (360) full 360 degree captures. Scene titles are formatted as: \textit{Scene Name (Number of Captured Frames in Input)}.   }
\label{fig:scenes}
\end{figure*}

\subsection{Data Collection} To record in-the-wild panorama video captures in unknown imaging conditions -- ranging from broad daylight to night-time photography -- we developed an Android-based data capture application, illustrated in Fig.~\ref{fig:app_layout}. The app records a stream of RAW images along with metadata, enabling us to leverage linear sensor data for noise-robust reconstruction. While there exist other RAW video and image recording apps, we found they were paid and closed-source, missing desired functionality (e.g., specifying ISO, exposure, and recording frame-rate), and/or failed to record desired data (e.g., gyroscope measurements). In contrast, our app records full-resolution full bit-depth RAW images at the hardware maximum of 30 frames per second, accelerometer and gyroscope measurements, and nearly all camera and image metadata exposed by the Android APIs -- a list of which is included in the supplementary material. We make this app available open-source at: {\color{URLBlue}\href{https://github.com/Ilya-Muromets/Pani}{github.com/Ilya-Muromets/Pani}}

We used a handheld Google Pixel 8 Pro cellphone to record a set of 50 scenes, a selection of which are presented in Fig.~\ref{fig:scenes}, which cover a wide span of both imaging settings and capture paths. These include traditional $360^\circ$ and $180^\circ$ panoramas, as well as linear horizontal and vertical pans, back-and-forth pans, and random-walk paths. We use the device's auto-exposure settings for recording, with sensor sensitivity varying from ISO $\approx20$ in daylight to ISO $\approx10,000$ for night-time scenes. Though we restrict exposure time to $\leq1/100$s to minimize motion blur during the relatively fast capture process (3-10 seconds depending on the length of the capture path). Recorded image sequences range between 30 and 100 frames depending, and include captures with the main (1x), ultrawide (0.5x), and telephoto (5x) cameras available on the device.

\begin{figure*}[h!]
 \centering
\includegraphics[width=\textwidth]{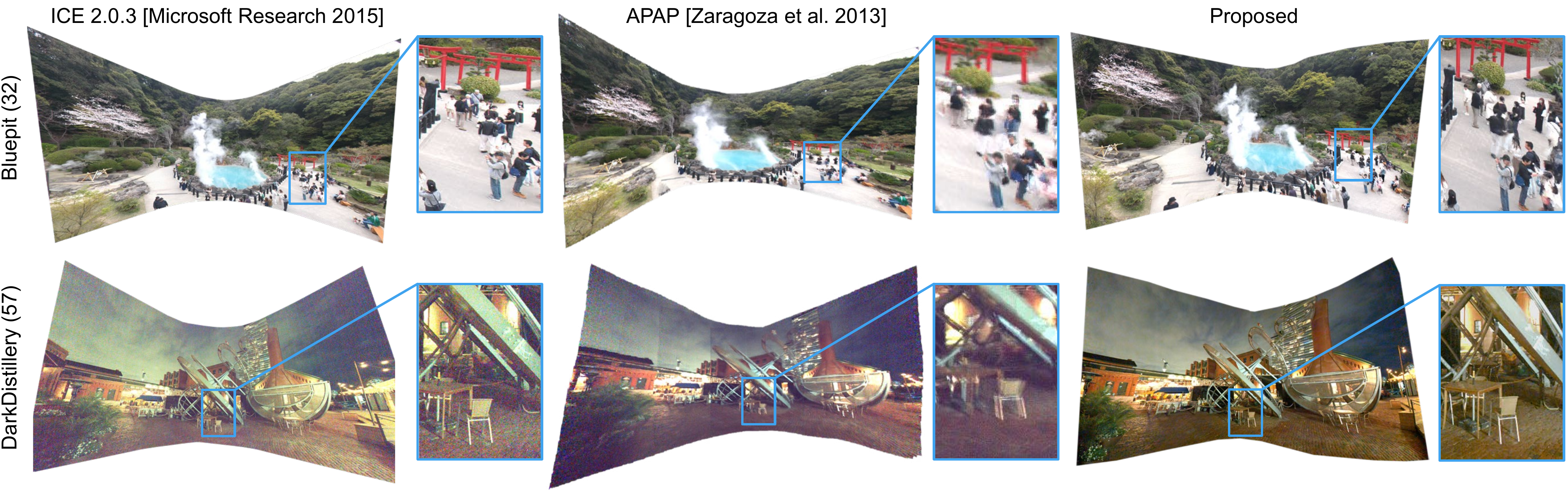}
  \caption{\textbf{Image Stitching Comparisons.} Visualizing rectilinear projections of the stitched panoramas, we see that APAP~\cite{zaragoza2013projective} averages multiple frames in \textit{DarkDistillery} to reduce noise, while ICE~\cite{MicrosoftICE} segments and freezes the motion of pedestrians in \textit{Bluepit}. Our proposed approach aims to do both, averaging multiple rays to reduce noise when possible while also preserving content in areas with local scene motion.
 }
\label{fig:stitching_comparison}
 \centering
\includegraphics[width=0.98\textwidth]{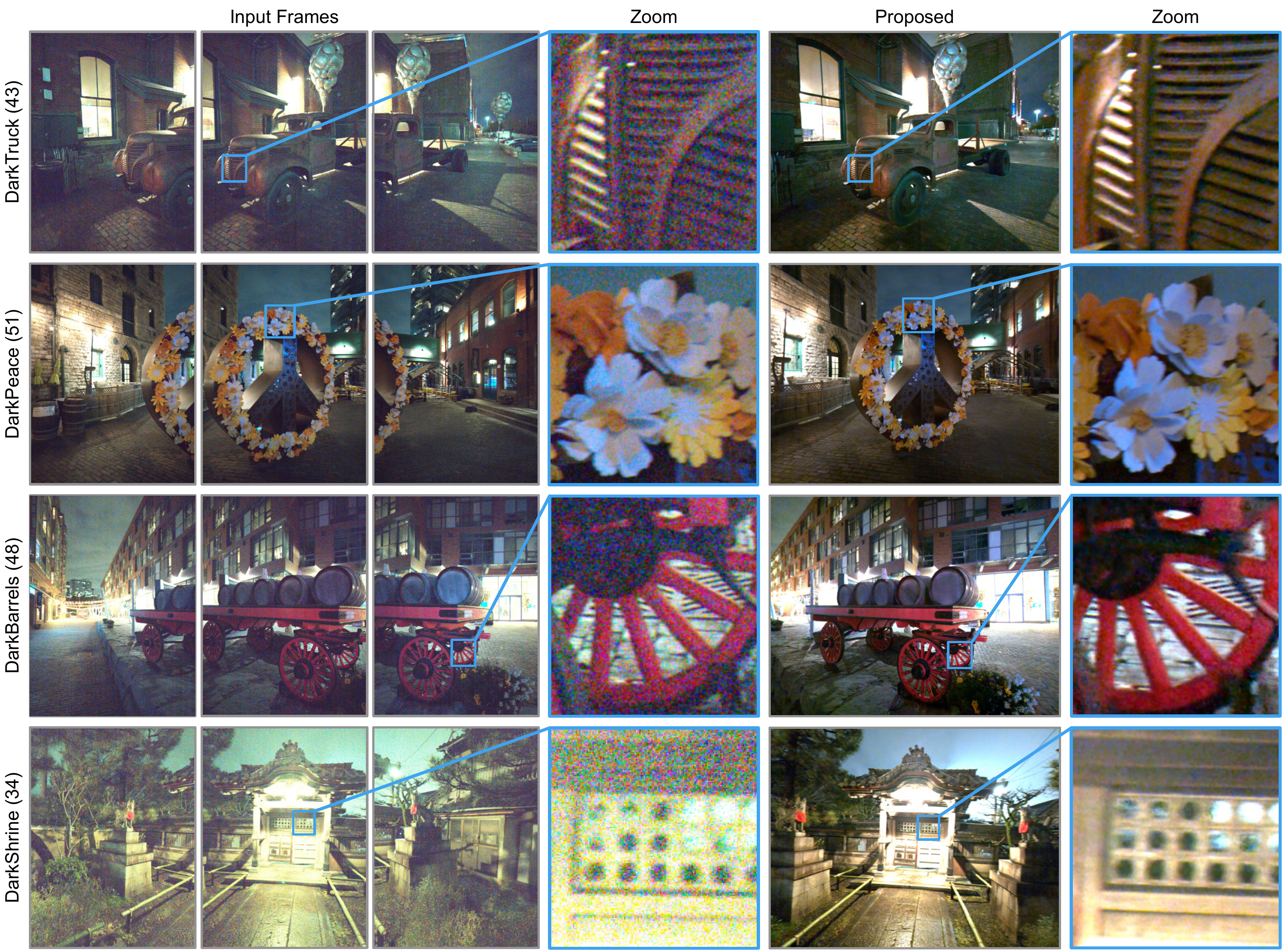}
  \caption{\textbf{Low-light Reconstruction.} Under low-light conditions, with sensor sensitivity at ISO 10,000 and exposure between $1/60\mathrm{s}$ and $1/120\mathrm{s}$, our proposed model is able to not only successfully reconstruct but also considerably denoise the captured scene. We recommend the reader to view the {\color{URLBlue}\href{https://light.princeton.edu/NeuLS}{associated video materials}} to see the effects of this denoising for interactive rendering.}
\label{fig:lowlight}
\end{figure*}

\begin{figure*}[h!]
 \centering
 
\includegraphics[width=\textwidth]{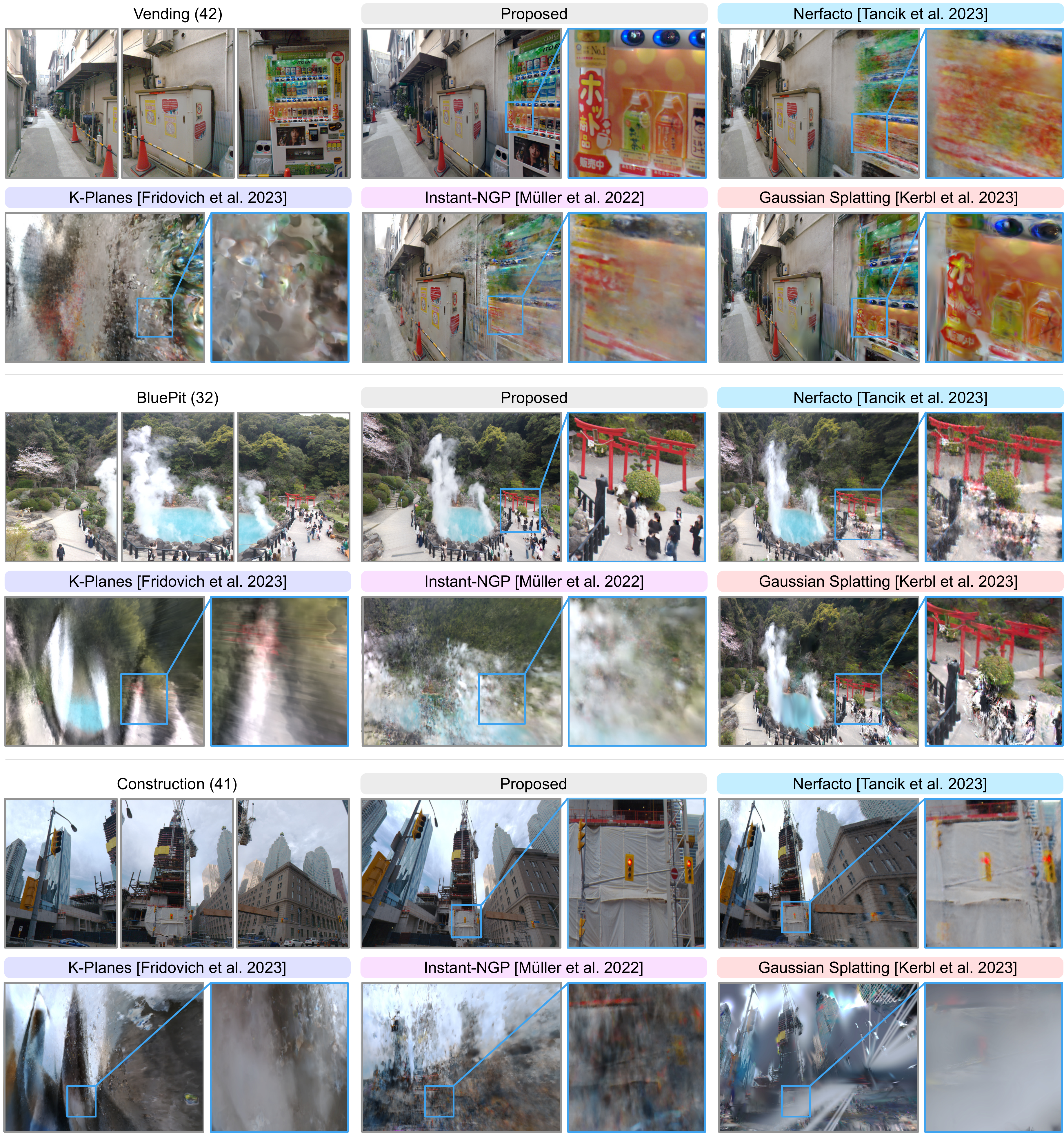}
  \caption{\textbf{Radiance Field Comparisons.} Compared to radiance field approaches, including other multi-resolution hash-based~\cite{tancik2023nerfstudio,muller2022instant} and non-volume-integrating~\cite{kerbl20233d,fridovich2023k} methods, we achieve significantly higher reconstruction quality over a range of settings. While Gaussian Splatting and Nerfacto are able to successfully overfit the center of most scenes (observed content), when the FOV is expanded to sample rays at wide angles they fail to correctly reconstruct fine images textures like the bottle labels in \textit{Vending}. In contrast, our neural light sphere model is able to reconstruct content in motion, like the pedestrians in \textit{BluePit} and fine parallax effects as in the traffic lights in \textit{Construction}. We recommend the reader to view the {\color{URLBlue}\href{https://light.princeton.edu/NeuLS}{associated video materials}} to better visualize these effects.  Scene titles are formatted as: \textit{Scene Name (Number of Captured Frames in Input)}.}
  \vspace{-1em}
\label{fig:nerf_comparison}
\end{figure*}

\subsection{Implementation Details} We implement our model in PyTorch with the tiny-cuda-nn framework~\cite{muller2021real}. It is trained via stochastic gradient descent with the Adam~\cite{kingma2014adam} optimizer ($\beta\,{=}\,[0.9,0.99]$, $\epsilon\,{=}\,10^{-9}$, weight decay $10^{-5}$, learning rate $10^{-3}$) for 100 epochs, with 200 batches of $2^{18}$ rays per epoch. Rotation weight $\eta_\textsc{r}\,{=}\,10^{-3}$. Networks ${h_\textsc{r}, h_\textsc{p}, h_\textsc{d}}$ are all identical 5 layer 128 hidden unit MLPs;  $h_\textsc{c}$ is single 32$\times$3 linear layer to discourage blending of view-dependent and static color. Encoding $\gamma_2$ is a 15-level hash grid, with grid resolutions spanning $4$ to $3145$ by powers of 1.61 for each encoded dimension, and with a backing table size of $2^{19}$. To constrain the spatial frequency of the view-dependent color and ray models, encoding $\gamma_1$ is a significantly lower-resolution grid, with 8 levels spanning resolutions of $4$ to $112$. Trained on a single Nvidia RTX 4090 GPU, our method takes approximately \textit{12 minutes} to fit a 40 frame 12-megapixel sequence, though we include results in the supplementary material for how this can be further accelerated to under 30 seconds for generating ``preview-quality'' reconstructions from 3-megapixel inputs. The image model takes 80 MB of disk space, and can render 1920$\times$1080px frames at 50 FPS. Critical to our core design goals discussed in Sec.~\ref{sec:neural_fields}, \textit{all parameters and training procedures are identical for all captures tested in all settings} (daytime, night-time, ultrawide, telephoto, etc.).

\section{Assessment}
In this section we compare our method to traditional image stitching and radiance field approaches. We then analyze the contributions of core model components, and confirm its applicability to the reconstruction of night-time scenes with noisy captures. For each scene we render views at 3x their original captured FOV, and include 3 input frames spanning the same FOV.
\subsection{Comparisons To Traditional Image Stitching}
While a large stitched image canvas is not the primary intended output of our neural light sphere model, as we focus on wide-view video rendering, comparisons to traditional image stitching methods help illustrate the challenges of this setting.

Presented in Fig.~\ref{fig:stitching_comparison}, we compare our approach to As-Projective-As-Possible (APAP) image stitching~\cite{zaragoza2013projective}, a robust parallax-tolerant cell-warping approach, and the Microsoft Image Composite Editor (ICE)~\cite{MicrosoftICE}, a polished software suite which performs globally projective warping and seam-blending to hide stitched image borders. APAP is able to warp and average multiple noisy measurements into a cleaner reconstruction, while ICE is restricted to stitching the borders of images together. However, ICE is significantly more resilient to motion-blur, freezing a sharp still frame of moving scene content. Our neural light sphere model offers both of these capabilities, averaging rays for better signal-to-noise ratio in static regions of the scene, while also more faithfully reconstructing dynamic content.

\subsection{Comparisons To Radiance Field Approaches} In Fig.~\ref{fig:nerf_comparison} we compare our hash-grid based, non-volume-sampling neural light sphere approach to several related radiance field methods including: \textit{K-Planes} \cite{fridovich2023k}, an explicit representation that also avoids volume sampling by representing the scene as a product of two-dimensional planar features; \textit{Gaussian Splatting} \cite{kerbl20233d}, which also avoids volume sampling through its forward-projection model; \textit{Instant-NGP} \cite{muller2022instant}, which makes use of the same multi-resolution hash-grid backing as our approach; and the \textit{Nerfacto}~\cite{tancik2023nerfstudio} model, a robust combined approach with a hash-grid backing and per-image appearance conditioning. Unfortunately, given the largely rotational motion of panorama captures, even using exhaustive feature matching both COLMAP~\cite{schonberger2016structure} and HLOC~\cite{sarlin2019coarse} failed to reconstruct poses for a significant portion of our tested scenes -- including virtually all telephoto and ultrawide captures. We thus limit the comparison scenes to ones where COLMAP produced valid poses, and enable camera pose optimization in baseline methods which support it. In contrast, we emphasize that, beyond selecting a directory to load from, \textit{there is no human interaction required between capture and reconstruction for our proposed pipeline}. 

Despite tuning feature grid and regularization parameters, we were unable to achieve high-quality reconstructions with \textit{K-Planes}, which appears to produce noisy low-dimensional approximations of the scene. We find that the other baseline methods tend to overfit input captures by placing content a large distance away from the estimated camera position, producing an effect similar to traditional image stitching~\cite{brown2007automatic}. We suspect this is in large part due to inaccurate initial camera pose estimates, which cause content to be incorrectly localized in 3D space, and cause the reconstructions to settle in geometrically inaccurate local-minima solutions. When the FOV of the simulated camera is expanded, and we simulate rays at steeper angles relative to the camera axis as compared to the input data, we see these overfitting artifacts as texture quality on the edges of the baseline renders significantly degrades. \textit{Instant-NGP} in particular struggles to extrapolate from data with low parallax or significant scene motion, such as the billowing steam clouds in \textit{Bluepit}. Conversely, our proposed approach is able to recover fine texture content in these areas, including readable text on the drink labels in \textit{Vending}.

\subsection{Applications to Low-light Photography} Illustrated in Fig.~\ref{fig:lowlight}, we find that, when trained on 10-bit linear RAW data, our neural light sphere model is robust to sensor noise as experienced in high ISO ($\geq 10,000$) settings during low-light photography. Similar to the findings of \cite{mildenhall2022nerf}, we find that by averaging rays that converge to identical scene points during training, our model learns a mean photometric solution for scene reconstruction, averaging out zero-mean Gaussian read noise. This also proves beneficial for non-light-limited settings, as we can lower exposure time for a single image to reduce motion blur during capture without risking failed reconstruction. Based on these initial findings, we expect a neural neural light sphere-style model could potentially be tailored for applications such as video denoising and astrophotography.

\section{Discussion and Future Work}
In this work we present a compact and robust neural light sphere model for handheld panoramic scene reconstruction. We demonstrate high-quality texture reconstruction in expanded field-of-view renders, with high tolerance to adverse imaging effects such as noise and localized pixel motion. 

\paragraph{Future Work} We hope that this work, and the accompanying metadata- and measurement-rich dataset, can encourage follow-on research into scene reconstruction under adverse imaging conditions. Many of the scenes, such as those illustrated in Fig.~\ref{fig:scenes}, purposely contain effects such as lens flare, snow, clouds, smog, reflections, sensor noise, and saturated high-dynamic range content. During in-the-wild data collection we found these effects unavoidable, highlighting the importance of robust reconstruction methods for practical computational photography. 

Beyond conventional photography, we believe this approach can be extended to industrial and scientific imaging settings such as satellite and telescope-based photography, scanning and array microscopes, and infrared or hyperspectral imaging. In particular, with a hardware-optimized hash-grid backing, our model design makes it computationally tractable to fit petapixel-and-larger data produced by these imaging modalities by breaking it into smaller ray batches -- e.g., a hash table size of $2^{22}$ reliably trains with batch size $2^{13}$ on a single Nvidia RTX 4090 GPU.

\begin{figure}[t!]
 \centering
\includegraphics[width=\linewidth]{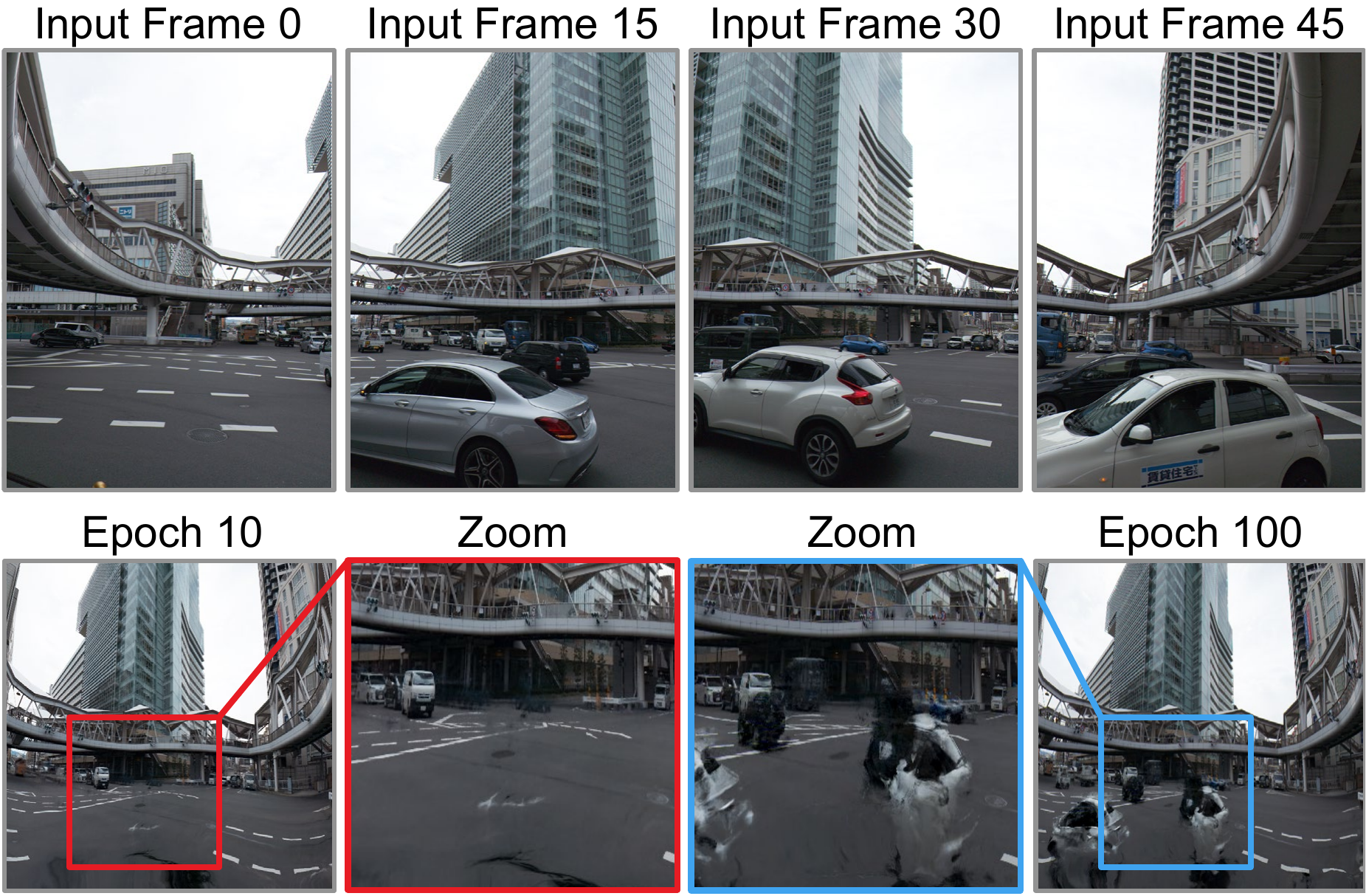}
  \caption{\textbf{Fast Occluders.} Objects such as bikes and cars, which quickly enter and exit the field-of-view of the camera, pose a challenge for scene reconstruction as they cannot be compactly modeled as a view-dependent effect. Shown in the example above, during early training the fast-moving cars are effectively erased from the reconstruction, which fits quickly to the median static pixel color. However, during later training stages, the view-dependent $h_\textsc{r}$ and $h_\textsc{d}$ models attempt (and fail) to reconstruct the content in motion, leading to transient car-shaped artifacts in the reconstruction.}
\label{fig:fast_cars}
\end{figure}

\paragraph{Limitations} Although the proposed method is robust to localized pixel motion and color changes -- e.g., swaying tree branches, flowing water -- it is not capable of reconstructing large fast-moving obstructions such as vehicles driving through a scene as shown in Fig.~\ref{fig:fast_cars}. This setting has posed a long-standing challenge for image-stitching and panoramic reconstruction works~\cite{szeliski2007image}, as when there are few observations of these occluders, this becomes a segmentation and tracking problem that is difficult to solve with a purely photometric approach such as ours. Similarly, without the ability to generate novel content, the camera path of the input capture strongly determines view synthesis performance -- e.g., a purely horizontal pan does not provide enough view information to simulate the effects of large vertical camera motion.

\begin{acks}
Ilya Chugunov was supported by an NSF GRFP (2039656). Felix Heide was supported by an Amazon Science Research Award, Packard Foundation Fellowship, Sloan Research Fellowship, Sony Young Faculty Award, the Project X Fund, and NSF CAREER (2047359). We thank Richard Szeliski, Jon Barron, and Ben Mildenhall for their valuable insights and discussions during this work's development.
\end{acks}


\addtocontents{toc}{\protect\setcounter{tocdepth}{1}}

\vspace{1em}
\noindent {\huge -- Supplementary Material --}

\setcounter{section}{0}
\renewcommand*{\theHsection}{chX.\the\value{section}}
\renewcommand\thesection{\Alph{section}}
\tableofcontents

\section{Implementation Details}
We compile a list of data recorded by our capture app and its uses in Tab.~\ref{table:metadata}. Our image processing pipeline follows the following sequence:
\begin{enumerate}
    \item Rearrange RAW data to BGGR format with \textit{color filter arrangement}
    \item Re-scale color channels as: \textit{(channel - black level)/(white level - black level)}
    \item Multiply by \textit{color correction gain}
    \item Multiply by inverse of \textit{shade map}
    \item Linearly interpolate gaps in mosaic (i.e., three interpolated values per red or blue, two interpolated values per green)
    \item Input into dataloader for training
\end{enumerate}
\vspace{1em}

\noindent To render final output images we then:
\begin{enumerate}
    \item Multiply RGB by the $3\times3$ \textit{color correction matrix}
    \item Re-scale color values with the \textit{tonemap curve}
\end{enumerate}
Or, optionally, skip this color correction to maximize render speed.

During training, we also use \textit{lens distortion} and \textit{rolling shutter skew} values to correct measurements on the ray level. Specifically we apply the lens distortion model as:

\begin{figure*}[h!]
 \centering
\includegraphics[width=\textwidth]{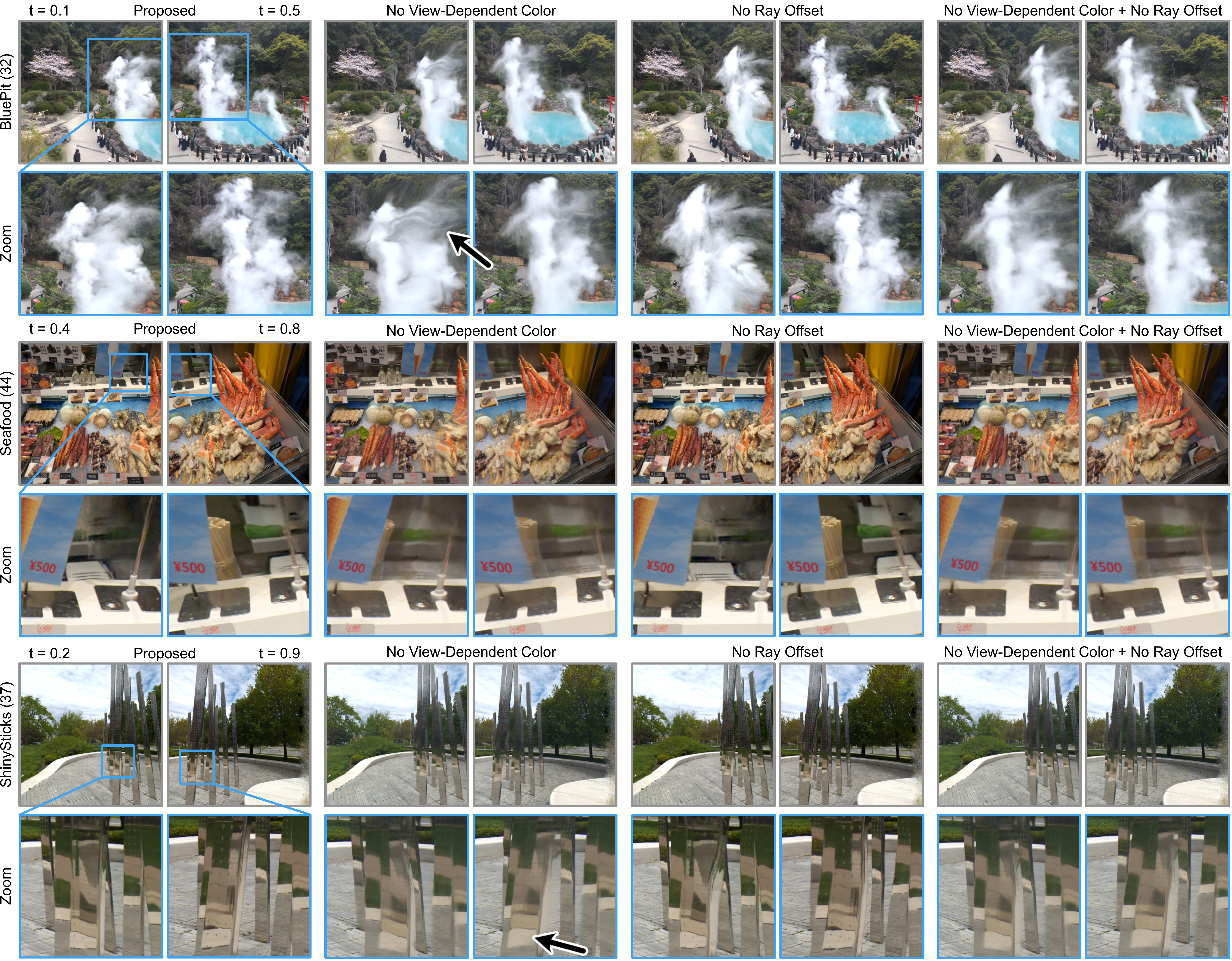}
 \caption{ \textbf{Model Component Analysis.} Shown above are the effects on reconstruction of zeroing out the contribution of the view-dependent color model $h_\textsc{d}\left(\gamma_1 (X);\, \theta_\textsc{d}\right)$, ray offset model $f_\textsc{r}(\hat{P}, X)$, or both models. We can observe that complex dynamic effects such as the steam clouds in \textit{BluePit} are produced by a combination of view-dependent color effects for the cloud texture, and ray offset for bulk motion. This is in contrast to the chopstick canister hidden behind the blue sign in \textit{Seafood}, which is almost entirely reconstructed with view-dependent color alone. In \textit{ShinySticks}, we observe how the sharp content and dots on the surface of the statue disappear when view-dependent color is removed, and large distortions in geometry appear when ray offset is omitted. }
\label{fig:modelcomponents}
\end{figure*}

\begin{equation}
\begin{aligned}
    x_{\text{dist}} &= x \left(1 + \kappa_1 r^2 + \kappa_2 r^4 + \kappa_3 r^6\right)
    + 2\kappa_4xy + \kappa_5 (r^2 + 2x^2) \\
    y_{\text{dist}} &= y \left(1 + \kappa_1 r^2 + \kappa_2 r^4 + \kappa_3 r^6\right)
    + 2\kappa_5 xy + \kappa_4 (r^2 + 2y^2)
\end{aligned}
\end{equation}

where $( r^2 = x^2 + y^2)$ is the squared radius from the optical center given by the camera \textit{intrinsics}. We also shift the time $n$ at which rays are sampled -- linearly interpolating translation \textit{T(n)} and rotation \textit{R(n)} -- by the row the ray was sampled from multiplied by the row rolling shutter delay given by \textit{rolling shutter skew}/\textit{image height}. We note that this rolling shutter delay had negligible effect on the overall reconstruction, possibly due to view-dependent ray offset model $f_\textsc{r}(\hat{P}, X)$ already able to compensate for it (introducing a row-dependent skew to the rays).

While we do not use data such as \textit{accelerometer values}, which give poor localization performance after double integration for pans, or \textit{ISO} and \textit{exposure time}, we hope that these may be of use in follow-on work. For example, while we keep exposure and ISO locked during our captures, it could be possible to combine bracketing~\cite{delbracio2021mobile} with panoramic capture to reconstruct ultra-HDR scenes.

\begin{table}[t!]
\centering
\begin{tabular}{@{} p{0.4\linewidth} p{0.5\linewidth} @{}}
\toprule
\textbf{Data} & \textbf{Purpose} \\
\midrule
intrinsics & ray projection ($K$) \\
color correction matrix & render output images \\
tonemap curve & render output images \\
shade map & correct RAWs (lens shading) \\
color filter arrangement & correct RAWs (BGGR) \\
lens distortion & correct RAWs (distortion) \\
color filter gains & correct RAW (color) \\ 
whitelevel & scale RAW data (max) \\
blacklevel & scale RAW data (min) \\
gyroscope values & rotation initialization ($G$) \\
timestamps & synchronize measurements \\
rolling shutter skew &  rolling shutter correction \\
accelerometer values & \textbf{unused} \\
ISO & \textbf{unused} \\
exposure time & \textbf{unused} \\
focus distance & \textbf{unused} \\
focal length & \textbf{unused} \\
lens extrinsics & \textbf{unused} \\
lens aperture & \textbf{unused} \\
neutral color point & \textbf{unused} \\
noise profile & \textbf{unused} \\

\bottomrule
\end{tabular}
\caption{\textbf{Recorded Data.} A non-exhaustive list of data, both used and unused in this project, recorded by our capture app.}
\label{table:metadata}
\end{table}

\begin{figure*}[h!]
 \centering
\includegraphics[width=\textwidth]{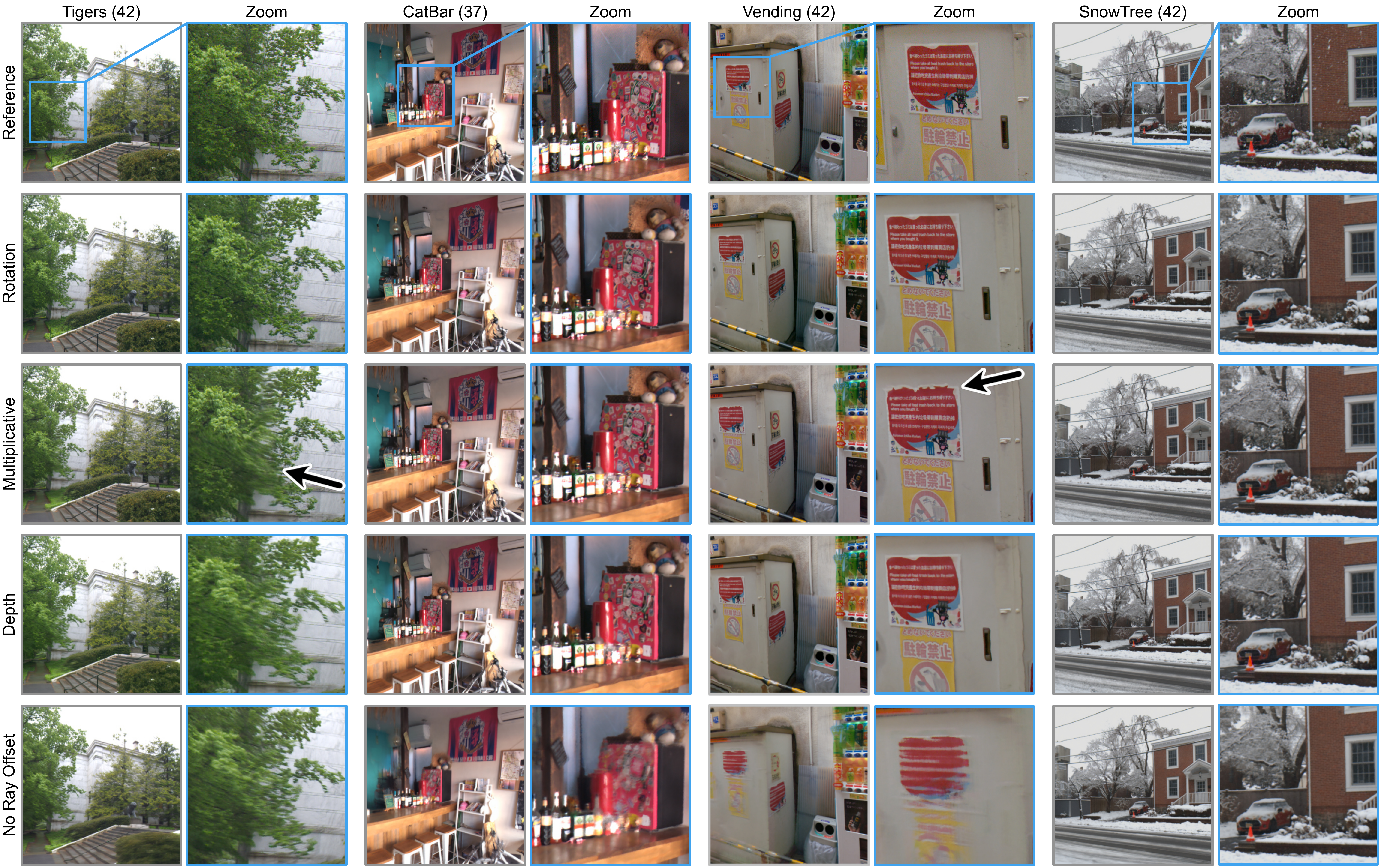}
  \caption{\textbf{Ray Offset Models.} Comparing scene reconstruction results for various ray offset models, it's clear from the \textit{No Ray Offset} results that many scenes such as \textit{CatBar} and \textit{Vending} contain significant parallax effects that a sphere projection model alone cannot compensate for. The \textit{Depth} and \textit{Multiplicative} models significantly improves reconstruction quality, albeit some regions in the \textit{Multiplicative} reconstructions suffer from distortions. The linearized \textit{Rotation} model avoids these artifacts while maintaining high reconstruction quality, recovering legible text in the \textit{Vending} scene.  }
\label{fig:offsets}
\end{figure*}

\begin{figure*}[h!]
 \centering
\includegraphics[width=\textwidth]{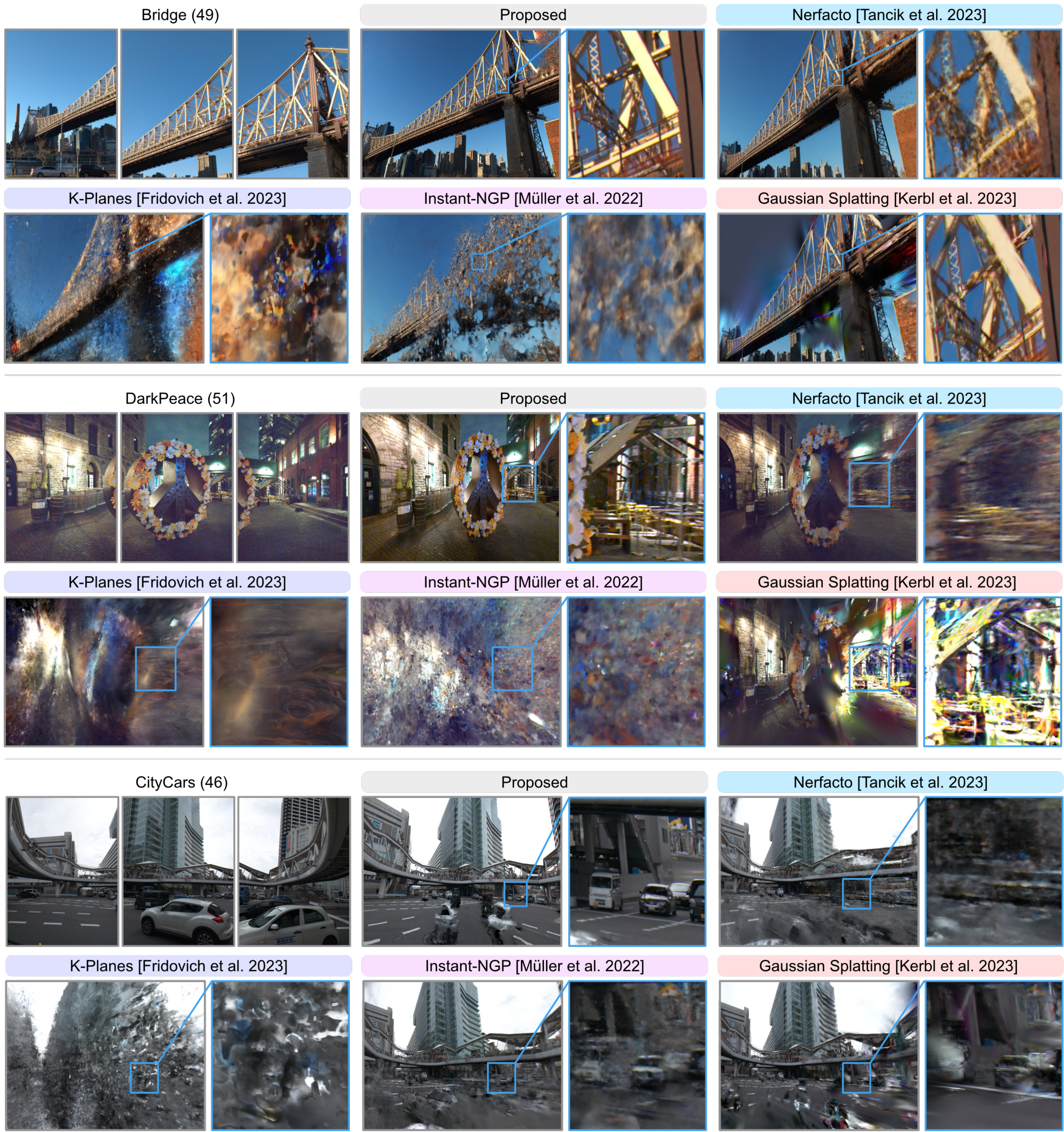}
  \caption{\textbf{Additional Radiance Field Comparisons.} Reconstruction results for a highly detailed back-and-forth \textit{Bridge} capture, night-time \textit{DarkPeace}, and \textit{CityCars} with fast-moving occluders. Scene titles are formatted as: \textit{Scene Name (Number of Captured Frames in Input)}}
\label{fig:nerf_comparison}
\vspace{2em}
\end{figure*}

\begin{figure*}[h!]
 \centering
\includegraphics[width=\textwidth]{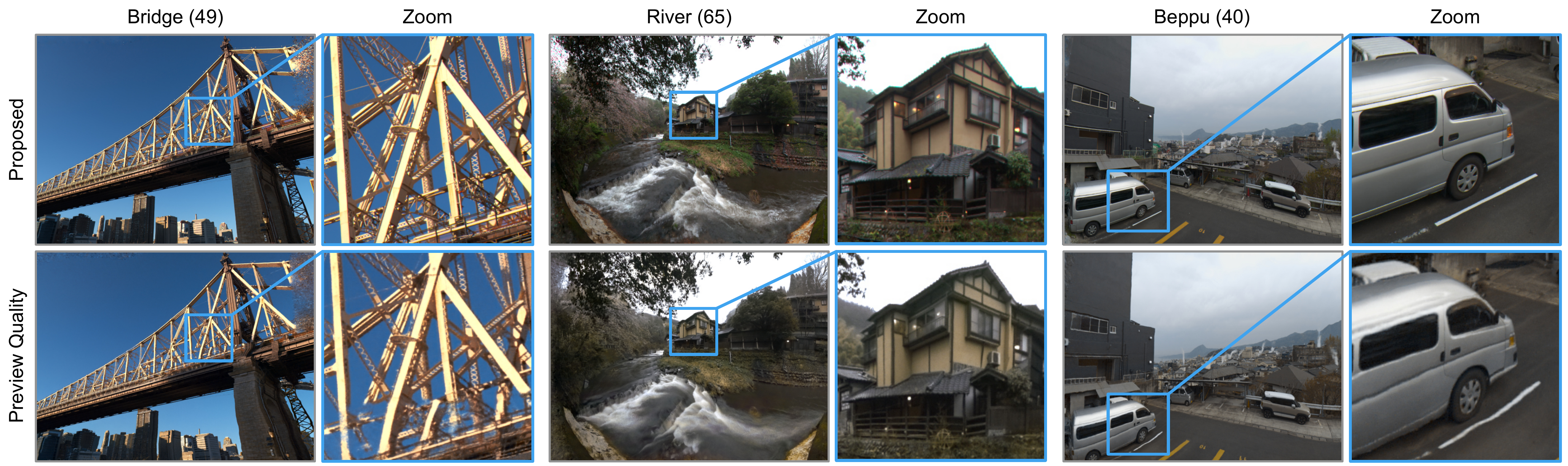}
  \caption{\textbf{Preview Quality Reconstructions.} Trained on 1/4 resolution inputs for 1/10th of the number of epochs, while they don't reach the full reconstruction quality of the proposed method, these ``Preview Quality'' reconstructions take less than 30 seconds of training time per scene.}
\label{preview}
\end{figure*}

\section{Model Component Analysis}
\label{sec:modelcomponent}
In Fig.~\ref{fig:modelcomponents} we visualize the independent contributions of the view-dependent color $f_{\textsc{c}}(\hat{P}^*,X)$ and ray offset models $f_\textsc{r}(\hat{P}, X)$ to our neural light sphere reconstructions. We train the model with both of these components active, and \textit{during inference time } we remove the output of the ray offset model
\begin{align}\label{eq:ray_distortion}
   \hat{D}^* =  D^*/\left\|D^*\right\|, \quad D^* = \cancelto{1}{\mathrm{rot}\left(\mathbf{R} = f_\textsc{r}(\hat{P}, X)\right)}\hat{D} = \hat{D},
\end{align}
remove the view-dependent color model
\begin{align}
    f_\textsc{c}(\hat{P}^*, X) &= h_\textsc{c}\left(h_\textsc{p}\left(\gamma_2 (\hat{P}^*); \,\theta_\textsc{p}\right) + \cancelto{0}{h_\textsc{d}\left(\gamma_1 (X);\, \theta_\textsc{d}\right)}; \, \theta_\textsc{c}\right)\nonumber \\
    f_\textsc{c}(\hat{P}^*, X) &= h_\textsc{c}\left(h_\textsc{p}\left(\gamma_2 (\hat{P}^*); \,\theta_\textsc{p}\right); \, \theta_\textsc{c}\right),
\end{align}
or remove both. From the resultant reconstructions, we can see how effects in the scene are modeled by one, both, or neither of these models. Static content on the surface of the sphere, such as the background folliage in \textit{BluePit} and \textit{ShinySticks} remains nearly identical in all reconstructions, which is entirely expected as this content exhibits almost no parallax and view-dependent color changes. In contrast, scene elements such as the reflections on the surface of \textit{ShinySticks} and the steam clouds in \textit{BluePit} require both the ray offset and view-dependent color models to work in tandem in order to produce these complex visual effects. This separability of our neural light sphere model also points towards a potentially interesting direction of future work, editing both content and its dynamics after reconstruction similar to a video mosaic~\cite{kasten2021layered} (e.g., turning the motion of the steam clouds into billowing smoke from a fire).
\section{Alternative Ray Offset Models} During the development of this work, we experimented with different ray offset models to model parallax and scene motion. This includes a \textit{Depth} model where we modify Eq. 4 of the main work to individually offset the radius of the sphere by  $f_\textsc{r}(\hat{P}, X)$ for each ray
\begin{align}\label{eq:sphere_intersection}
   \hat{P}^* &=  P/\left\|P\right\|, \quad P \,{=}\, \left[\arraycolsep=2.0pt
    \begin{array}{c}
    P_x \\
    P_y \\
    P_z \\
    \end{array}\right] = O + t\hat{D}\nonumber\\
    t &= -\left(O \cdot \hat{D}\right) + \sqrt{(O \cdot \hat{D})^2 - (\left\| O\right\|^2 - (1+f_\textsc{r}(\hat{P}, X)))},
\end{align}
simulating a depth map stretched across the inside surface of the sphere model. Another model we tested was a \textit{Multiplicative} ray offset
\begin{align}\label{eq:ray_distortion}
   \hat{D}^* =  D^*/\left\|D^*\right\|, \quad D^* = (1 + f_\textsc{r}(\hat{P}, X))\circ\hat{D},
\end{align}
where $\circ$ denotes element-wise multiplication. In Fig.~\ref{fig:offsets} we can see how this model further sharpens content when compared to the \textit{Depth} model, but leads to blur and distortions in the scene where a large multiplicative offset causes rays to be ``pushed" out of a region in the scene. The final ray offset model we chose was a linearized \textit{Rotation} model
\begin{align}\label{eq:ray_distortion}
   \hat{D}^* =  D^*/\left\|D^*\right\|, \quad D^* = \mathrm{rot}\left(\mathbf{R} = f_\textsc{r}(\hat{P}, X)\right)\hat{D},
\end{align}
which we observed to lead to high reconstruction quality without the distortions observed in the \textit{Multiplicative} model. Here a larger $f_\textsc{r}(\hat{P}, X)$ rotates a region of rays together a larger distance, rather than pushing them out of a region on the sphere.

To compare these models, we remove the view-dependent color model $h_\textsc{d}\left(\gamma_1 (X);\, \theta_\textsc{d}\right)$ as outlined in Sec.~\ref{sec:modelcomponent} \textit{during training}, not just during inference. As otherwise this $h_\textsc{d}\left(\gamma_1 (X);\, \theta_\textsc{d}\right)$ can compensate for content that was not correctly reconstructed by the ray offset model. We compare reconstruction results for these offset models in Fig.~\ref{fig:offsets}, noting that for scenes such as \textit{Vending} and \textit{CatBar} with large amount of parallax the choice of offset model significantly affects reconstruction quality. Conversely, for \textit{SnowTree}, where content is far from the camera, all models produce similar reconstructions, emphasizing the importance of collecting a diverse set of scenes to holistically evaluate in-the-wild image stitching.

\section{Additional Reconstruction Results}
In Fig.~\ref{fig:nerf_comparison} we showcase additional reconstruction results and comparisons to radiance field baselines: \textit{K-Planes} \cite{fridovich2023k}, \textit{Gaussian Splatting} \cite{kerbl20233d},  \textit{Instant-NGP} \cite{muller2022instant}, and \textit{Nerfacto}~\cite{tancik2023nerfstudio}. Noteably, we see in \textit{Bridge} the high resolution reconstruction enabled by our method, which is able to correctly resolve the cross-hatch bars in the bridge's support structure. In \textit{DarkPeace} we see that while \textit{Nerfacto} and \textit{Gaussian Splatting} successfully reconstruct the left side of the scene, the area of maximum overlap where the capture started, they produce extremely noisy reconstructions at the end of the capture sequence, with \textit{Instant-NGP} failing to reconstruct any of the scene. In \textit{CityCars} we can observe how, while our neural light sphere model is not able to reconstruct the fast-moving cars, the reconstruction artifacts only disrupt local content. Zooming into the background, we can still resolve the static cars, unlike the baseline methods, which produce reconstructions corrupted by motion artifacts.

\section{Preview-Scale Rendering}
While the reconstructions shown in the main text are relatively fast to train compared to the average neural radiance field approach, during model exploration and development we found it extremely beneficial to be able to quickly test large collections of scenes. By down-sampling the input data from full resolution 12-megapixel images to 1/4 resolution 3-megapixel images, dividing the max epochs by 10, and removing tensorboarding operations we are able to render ``preview quality'' scenes in less than 30 seconds. While there is a notable drop in quality for some scene content, as seen in the deformation of the grey car in the \textit{Beppu} example shown in Fig.~\ref{preview}, other scenes reach high reconstruction quality even in this short training time. Even zooming into the \textit{River} scene it is difficult to see a change in quality between the two reconstructions; suggesting that with some training augmentation, near-instant reconstruction could be possible for some subset of panoramic video captures.

\addtocontents{toc}{\protect\setcounter{tocdepth}{0}}
\bibliographystyle{ACM-Reference-Format}
\bibliography{main}

\end{document}